# Modeling Discrete Interventional Data using Directed Cyclic Graphical Models

**Mark Schmidt and Kevin Murphy**
Department of Computer Science
University of British Columbia
{schmidtm,murphyk}@cs.ubc.ca

## Abstract

We outline a representation for discrete multivariate distributions in terms of interventional potential functions that are globally normalized. This representation can be used to model the effects of interventions, and the independence properties encoded in this model can be represented as a directed graph that allows cycles. In addition to discussing inference and sampling with this representation, we give an exponential family parametrization that allows parameter estimation to be stated as a convex optimization problem; we also give a convex relaxation of the task of simultaneous parameter and structure learning using group $\ell_1$-regularization. The model is evaluated on simulated data and intracellular flow cytometry data.

## 1 Introduction

Graphical models provide a convenient framework for representing independence properties of multivariate distributions (Lauritzen, 1996). There has been substantial recent interest in using graphical models to model data with interventions, that is, data where some of the variables are set experimentally. Directed acyclic graphical (DAG) models represent a joint distribution over variables as a product of conditional probability functions, and are a convenient framework for modeling interventional data using Pearl's *do*-calculus (Pearl, 2000). However, the assumption of acyclicity is often inappropriate; many models of biological networks contain feedback cycles (for example, see Sachs et al. (2005)). In contrast, undirected graphical models represent a joint distribution over variables as a globally normalized product of (unnormalized) clique potential functions, allowing cycles in the undirected graph. However, the symmetry present in undirected models means that there is no natural notion of an intervention: For undirected models, there is no difference between observing a variable ('seeing') and setting it by intervention ('doing').

Motivated by the problem of using cyclic models for interventional data, in this paper we examine a class of directed cyclic graphical models that represent a discrete joint distribution as a globally normalized product of (unnormalized) *interventional* potential functions, leading to a convenient framework for building cyclic models of interventional data. In §2, we review several highlights of the substantial literature on directed cyclic graphical models and the closely related topic of representing distributions in terms of conditional functions. Subsequently, we discuss representing a joint distribution over discrete variables with interventional potential functions (§3), the Markov independence properties resulting from a graphical interpretation of these potentials (§4), modeling the effects of interventions under this representation (§5), interpreting the model and interventions in the model in terms of a data generating process involving feedback (§6), inference and sampling in the model (§7), parameter estimation with an exponential family representation (§8), and a convex relaxation of structure learning (§9). Our experimental results (§10) indicate that this model offers an improvement in performance over both directed and undirected models on both simulated data and the data analyzed in (Sachs et al., 2005).

## 2 Related Work

Our work is closely related to a variety of previous methods that express joint distributions in terms of conditional distributions. For example, the classic work on pseudo-likelihood for parameter estimation (Besag, 1975) considers optimizing the set of conditional distributions as a surrogate to optimizing the joint distribution. Heckerman et al. (2000) have advocated the advantages of dependency networks, directed cyclic models expressed in terms of conditional probability distributions (where 'pseudo-Gibbs' sampling is used to answer probabilistic queries). They argue that the set of conditional distributions may be simpler to specify than a joint distribution, and can be computationally cheaper to fit. Closely related to dependency networks is the work of Hofmann and Tresp (1997), as well as work on conditionally specified distributions (Arnold et al., 2001; Heckerman et al., 2004) (and the references contained in these works). However, to our knowledge previous work on these models has not considered using globally normalized 'conditional' potential functions and trying to optimize the joint distribution defined by their product, nor has it considered modeling the effects of interventions.

Our work is also closely related to work on path diagrams and structural equation models (SEMs) (Wright, 1921), models of functional dependence that have long been used in genetics, econometrics, and the social sciences (see Pearl (2000)). Spirtes (1995) discusses various aspects of 'non-recursive' SEMs, which can be used to represent directed cyclic feedback processes (as opposed to 'recursive' SEMs that can be represented as a DAG). Spirtes (1995) shows that d-separation is a valid criterion for determining independencies from the graph structure in linear SEMs. Pearl and Dechter (1996) prove an analogous result that d-separation is valid for feedback systems involving discrete variables. Modeling the effects of interventions in SEMs is discussed in (Strotz and Wold, 1960). Richardson (1996a,b) examines the problem of deciding Markov equivalence of directed cyclic graphical models, and proposes a method to find the structure of directed cyclic graphs. Lacerda et al. (2008) recently proposed a new method of learning cyclic SEMs for certain types of (non-interventional) continuous data. The representation described in this paper is distinct from this prior work on directed cyclic models in that the Markov properties are given by moralization of the directed cyclic graph (§4), rather than d-separation. Further, we use potential functions to define a joint distribution over the variables, while SEMs use deterministic functions to define the value of a child given its parents (and error term).

A third thread of research related to this work is prior work on combinations of directed and undirected models. Modeling the effects of interventions in chain graphs is thoroughly discussed in Lauritzen and Richardson (2002). Chain graphs are associated with yet another set of Markov properties and, unlike non-recursive SEMs and our representation, require the restriction that the graph contains no partially directed cycles. Also closely related are directed factor graphs (Frey, 2003), but no interventional semantics have been defined for these models.

## 3 Interventional Potential Representation

We represent the joint distribution over a set of discrete variables $x_i$ (for $i \in \{1,\ldots,n\}$) as a globally normalized product of non-negative interventional potential functions

$$p(x_1,\ldots,x_n) = \frac{1}{Z}\prod_{i=1}^{n}\phi(x_i|x_{\pi(i)}),$$

where $\pi(i)$ is the set of 'parents' of node $i$, and the function $\phi(x_i|x_{\pi(i)})$ assigns a non-negative potential to each joint configuration of $x_i$ and its parents $x_{\pi(i)}$. The normalizing constant

$$Z = \sum_{\vec{x}}\prod_{i}\phi(x_i|x_{\pi(i)}),$$

enforces that the sum over all possible configurations of $\vec{x}$ is unity. In contrast, undirected graphical models represent the joint distribution as a globally normalized product of non-negative potential functions defined on a set of $C$ cliques,

$$p(x_1,\ldots,x_n) = \frac{1}{Z}\prod_{c=1}^{C}\phi(x_c).$$

While in undirected graphical models we visualize the structure in the model as an undirected graph with edges between variables in the same cliques, in the interventional potential representation we can visualize the structure of the model as a directed graph $G$, where $G$ contains a directed edge going into each node from each of its parents. The global normalization allows the graph $G$ defining these parent-child relationships to be an arbitrary directed graph between the nodes.

We obtain DAG models in the special case where the graph $G$ is acyclic and for each node $i$ the potentials satisfy the local normalization constraint $\forall_{x_{\pi(i)}}\sum_{x_i}\phi(x_i|x_{\pi(i)}) = 1$. With these restrictions, the interventional potentials represent conditional probabilities, and it can be shown that $Z$ is constrained to be 1[1]. However, unlike DAG models, in our new representation the potentials do not need to satisfy any local normalization conditions, $p(x_i|x_{\pi(i)})$ will not generally be proportional to $\phi(x_i|x_{\pi(i)})$, and $G$ is allowed to have directed cycles.

[1]Because the global normalization makes the distribution invariant to re-scaling of the potentials, the distribution will also be equivalent to a DAG model under the weaker condition that the graph is acyclic and for each node $i$ there exists a constant $c_i$ such that $\forall_{x_{\pi(i)}}\sum_{x_i}\phi(x_i|x_{\pi(i)}) = c_i$. The conditional probability functions in the corresponding DAG model are obtained by dividing each potential by the appropriate $c_i$.

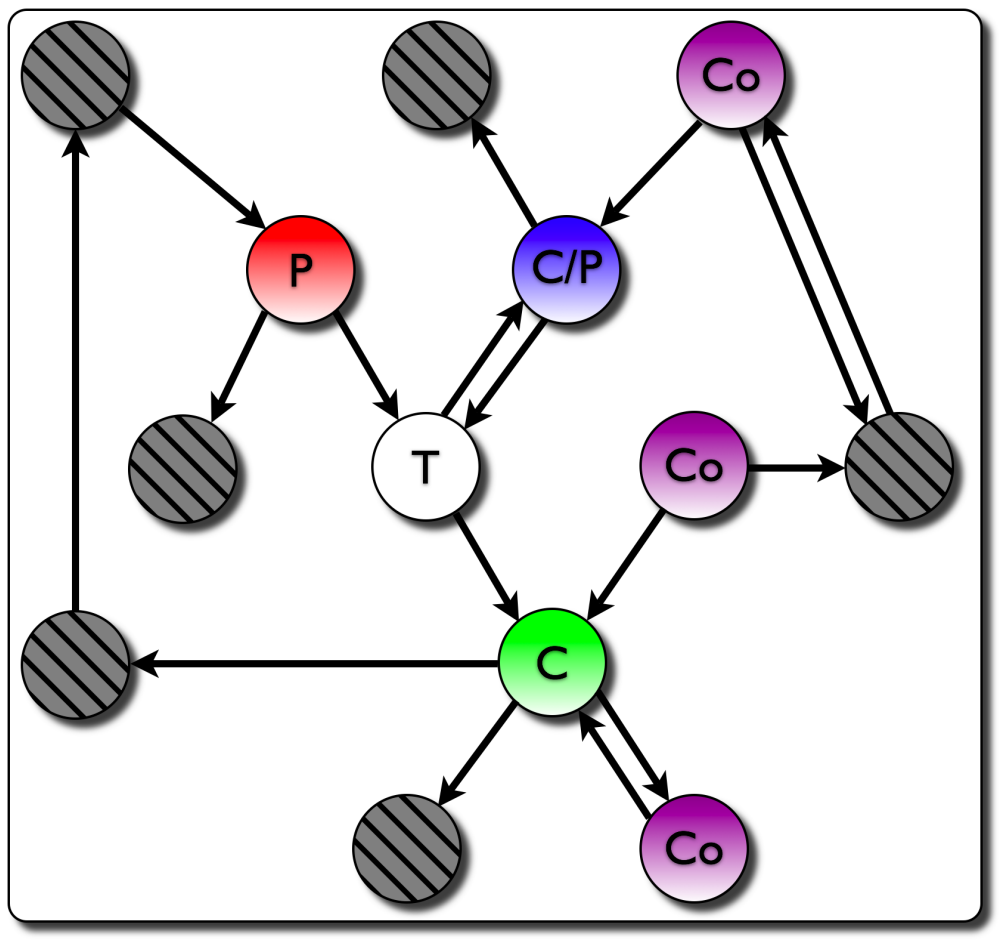

Figure 1: The Markov blanket for node (T) includes its parents (P), children (C), and co-parents (Co). The node labeled C/P is both a child and a parent of T, and together they form a directed 2-cycle.

## 4 Markov Independence Properties

We define a node's Markov blanket to be its parents, children, and co-parents (other parents of the node's children). If the potential functions are strictly positive, then each node in the graph is independent of all other nodes given its Markov blanket:

$$p(x_i|x_{-i}) = \frac{p(x_i, x_{-i})}{\sum_{x_i'} p(x_i', x_{-i})}$$
$$= \frac{\frac{1}{Z}\phi(x_i|x_{\pi(i)})\prod_{j\neq i, i\notin\pi(j)}\phi(x_j|x_{\pi(j)})\prod_{j\neq i, i\in\pi(j)}\phi(x_j|x_{\pi(j)})}{\sum_{x_i'}\frac{1}{Z}\phi(x_i'|x_{\pi(i)})\prod_{j\neq i, i\notin\pi(j)}\phi(x_j|x_{\pi(j)})\prod_{j\neq i, i\in\pi(j)}\phi(x_j|x_i', x_{\pi(j)\backslash i})}$$
$$= \frac{\phi(x_i|x_{\pi(i)})\prod_{j\neq i, i\in\pi(j)}\phi(x_j|x_{\pi(j)})}{\sum_{x_i'}\phi(x_i'|x_{\pi(i)})\prod_{j\neq i, i\in\pi(j)}\phi(x_j|x_i', x_{\pi(j)\backslash i})} = p(x_i|x_{MB(i)}).$$

Above we have used $x_{-i}$ to denote all nodes except node $i$, and $x_{MB(i)}$ to denote the nodes in the Markov blanket of node $i$. Figure 1 illustrates an example of a node's Markov blanket.

In addition to this local Markov property, we can also use graphical operations to answer arbitrary queries about (conditional) independencies in the distribution. To do this, we first form an undirected graph by (i) placing an undirected edge between all co-parents that are not directly connected, (ii) replacing all directed 2-cycles with a single undirected edge, and (iii) replacing all remaining directed edges with undirected edges. To test whether a set of nodes $P$ is independent of another set $Q$ conditioned on a set $R$ (denoted $P \perp Q|R$), it is sufficient to test whether a path exists between a node in $P$ and a node in $Q$ that does not pass through any nodes in $R$. If no such path exists, then the factorization of the joint distribution implies $P \perp Q|R$. This procedure is closely related to the separation criterion for determining independencies in undirected graphical models (see Koller and Friedman (2009)), and it follows from a similar argument that this test for independence is sound[2].

## 5 Effects of Interventions

Up to this point, the directed cyclic model can be viewed as a re-parameterization of an undirected model. In this section we consider interventional data, which is naturally modeled using the interventional potential representation, but is not naturally modeled by the (symmetric) clique potentials used in undirected graphical models.

In DAG models, we can incorporate an *observation* that a variable $x_i$ takes on a specific value using the rules of conditional probability (eg. $p(x_{2:n}|x_1) = p(x_{1:n})/p(x_1)$). To model the effect of an *intervention*, where a variable $x_i$ is explicitly forced to take on a specific value, we first remove the conditional mass function $p(x_i|x_{\pi(i)})$ from the joint probability, and then use the rules of conditional probability on the resulting modified distribution (Pearl, 2000). Viewed graphically, removing the term from the joint distribution deletes the edges going into the target of intervention, but preserves edges going out of the target.

By working with the interventional potential representation, we can define the effect of setting a variable by intervention analogously to DAG models. Specifically, setting a node $x_i$ by intervention corresponds to removing the potential function $\phi(x_i|x_{\pi(i)})$ from the joint distribution. The corresponding graphical operation is similar to DAG models, in that edges going into targets of interventions are removed while edges leaving targets are left intact[3]. In the case of directed 2-cycles, the effect of an intervention is thus to change the directed 2-cycle into a directed edge away from the target of intervention. Figure 2 illustrates the effects of interventions in the case of a single edge, and in the case of a directed 2-cycle. The independence properties of the interventional distribution can be determined graphically in the same way as the observational (non-interventional) distribution, by working with the modified graph. Note that intervention can not only affect the independence properties between parent and child nodes, but also between co-parents of the node set by intervention. Figure 3 gives an example.

[2]For a specific set of interventional potential functions, there may be additional independence properties that are not encoded in the graph structure (for example, if we have deterministic dependencies or if we enforce the local normalization conditions needed for equivalence with DAG models). However, these are a result of the exact potentials used and will disappear under minor perturbations of their values.

[3]The effect of interventions for SEMs is also analogous, in that an intervention replaces the structural equation for

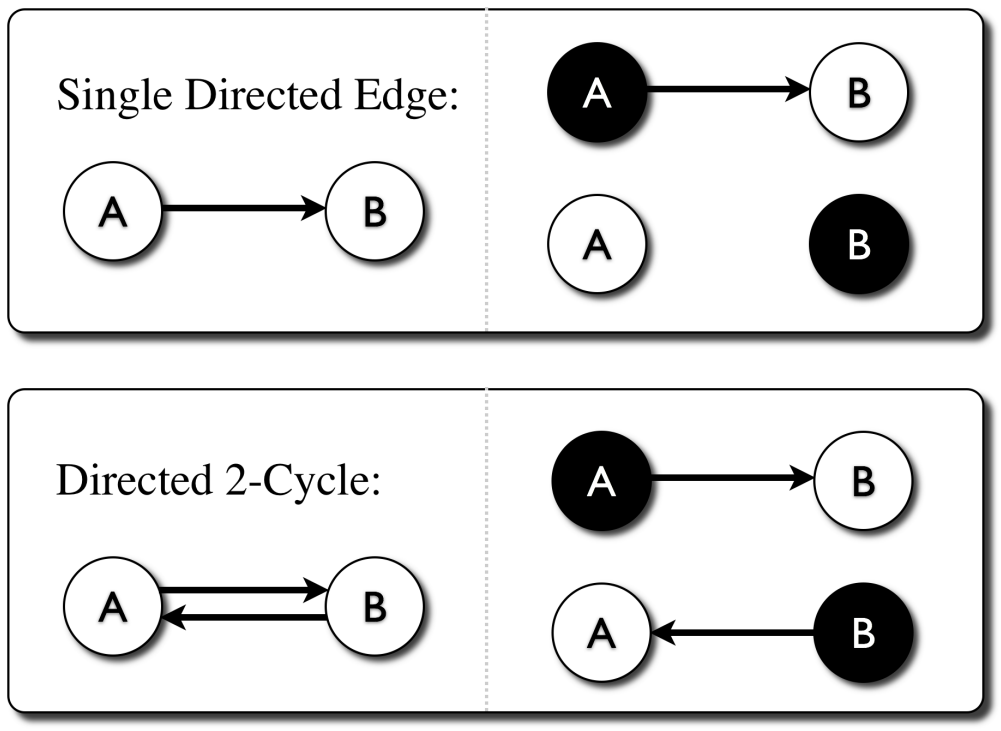

Figure 2: Effects of interventions for a single directed edge (top) and a directed 2-cycle (bottom). On the left side we show the unmodified graph structure. On the right side, we show the effect on the graph structure of intervening on the shaded node. For a single edge, the intervention leaves the graph unchanged if the parent is the target, and severs the edge if the child is the target. For the directed 2-cycle, we are left with a single edge leaving the target.

These interventional semantics distinguish the interventional potential representation of an undirected model from the clique potential representation. Intuitively, we can think of the directions in the graph as representing undirected influences that are robust to intervention on the parent but not the child. That is, a directed edge from node $i$ to $j$ represents an undirected statistical dependency that remains after intervention on node $i$, but would not exist after intervention on node $j$.

## 6 A Data Generating Process

Following §6 of Lauritzen and Richardson (2002), we can consider a Markov chain Monte Carlo method for simulating from a distribution represented with interventional potentials. In particular, consider the random Gibbs sampler where, beginning from some initial $\vec{x}^0$, at each iteration we choose a node $i$ at random and sample $x_i$ according to $p(x_i|x_{MB(i)})$. If we stop this algorithm after a sufficiently large number of iterations that the Markov chain was able to converge to its stationary (equilibrium) distribution, then the final value of $\vec{x}$ represents a sample from the distribution. This data generating process involves feedback in the sense that the value of each node is affected by all nodes connected (directly or indirectly) to it in the graph. However, this process is different than previous cyclic feedback models in that the instantaneous value of a variable is determined by its entire Markov blanket (as in undirected models), rather than its parents alone (as in directed models).

The data generating process under conditioning (by observation) sets the appropriate values of $\vec{x}^0$ to their observed values, and does not consider selecting observed nodes in the random update. The data generating process under intervention similarly excludes updating of nodes set by intervention. However, interventions can also affect the updating of nodes not set by intervention, since (i) a child set by intervention may be removed from the Markov blanket, and/or (ii) co-parents of a child set by intervention may be removed from the Markov blanket. From this perspective, we can give an interpretation to interventions in the model; an intervention on a node $i$ will remove or modify (in the case of a directed 2-cycle) the instantaneous statistical dependencies between node $i$ and its parents (and between co-parents of node $i$) in the equilibrium distribution. Of course, even if instantaneous statistical dependencies are removed between a parent and child in the equilibrium distribution, the child set by intervention may still be able to indirectly affect its the target of intervention with a simple assignment operator (Strotz and Wold, 1960). Graphically, this modifies the path diagram so that edges going into the target of intervention are removed but outgoing edges are preserved.

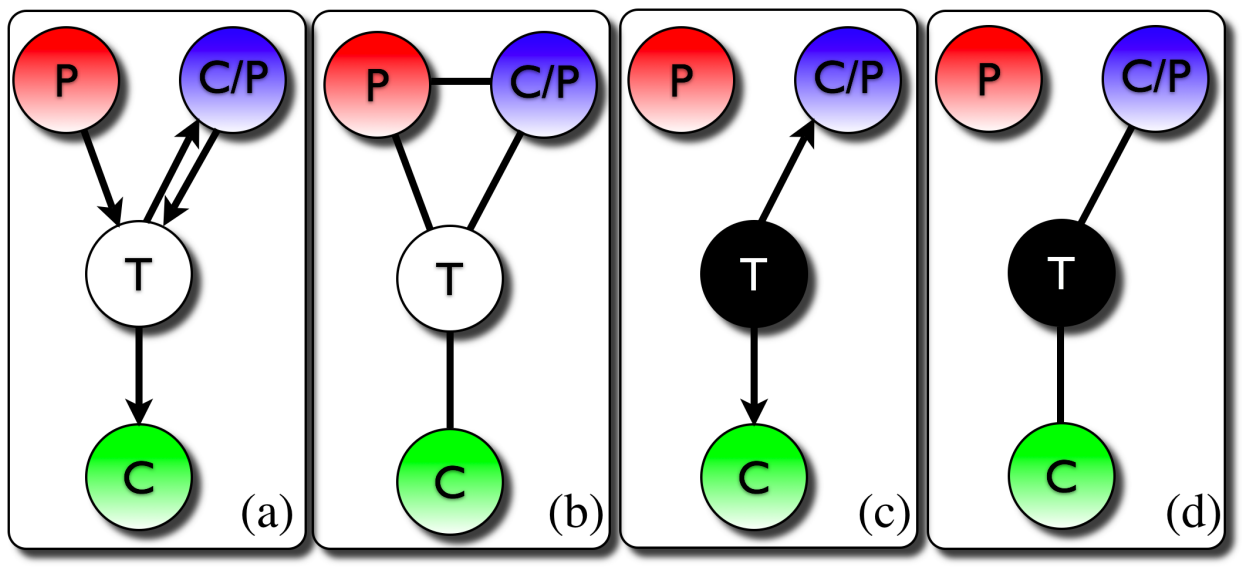

Figure 3: Independence properties in a simple graph before and after intervention on node T. From left to right we have (a) the original graph, (b) the undirected graph representing independence properties in the observational distribution, (c) the modified graph after intervention on T, and (d) the undirected graph representing independence properties in the interventional distribution.

parent in the equilibrium distribution if other paths exist between the child and parent in the graph.

## 7 Inference and Sampling

When the total number of possible states is small, inference in the model can be carried out in a straightforward way. Computing node marginals involves summing the potentials for particular configurations and dividing by the normalizing constant

$$p(x_i = c) = \frac{1}{Z} \sum_{\vec{x}} \mathbb{I}_c(x_i) \prod_j \phi(x_j | x_{\pi(j)}),$$

where $\mathbb{I}_c(x_i)$ is the indicator function, taking a value of 1 when $x_i$ takes the state $c$ and 0 otherwise.

Computing marginals over the configurations of several nodes is performed similarly, while inference with observations can be computed using the rules of conditional probability. We model interventions by removing the appropriate interventional potential function(s) and computing a modified normalizing constant $Z'$; the rules of conditional probability are then used on the modified distribution. For example, if we intervene on node $k$, we can compute the marginal of a node $i$ (for $i \neq k$) using

$$p(x_i = c | do(x_k)) = \frac{1}{Z'} \sum_{x_{-k}} \mathbb{I}_c(x_i) \prod_{j \neq k} \phi(x_j | x_{\pi(j)}),$$

where the modified normalizing constant is

$$Z' = \sum_{x_{-k}} \prod_{j \neq k} \phi(x_j | x_{\pi(j)}).$$

We can generate samples from the model using an inverse cumulative distribution function method; we generating a uniform deviate $\mathcal{U}$ in $[0, 1]$, then compute each term in the sum $\sum_{\vec{x}} \frac{1}{Z} \prod_i \phi(x_i | x_{\pi(i)})$ and stop at the configuration where this sum first equals or surpasses $\mathcal{U}$.

For larger graphs, these computations are intractable due to the need to compute the normalizing constant (and other sums over the set of possible configurations). Fortunately, the local Markov property will often make it trivial to implement Gibbs sampling (or block-Gibbs sampling) methods for the model (Geman and Geman, 1984)[4]. It is also possible to take advantage of dynamic programming methods for exact inference (when the graph structure permits), and more sophisticated variational and stochastic inference methods (for example, see Koller and Friedman (2009)). However, a discussion of these methods is outside the scope of this paper.

## 8 Exponential Family Parameter Estimation

For a fixed graph structure, the maximum likelihood estimate of the parameters given a data matrix $X$ (containing $m$ rows where each row is a sample of the $n$ variables) can be written as the minimization of the negative log-likelihood function in terms of the parameters $\theta$ of the interventional potentials.

$$\begin{aligned} -\log p(X|\theta) = &- \sum_{d=1}^{m} \sum_{i=1}^{n} \log(\phi(X_{d,i} | X_{d,\pi(i)}, \theta)) \\ &+ m \log Z(\theta). \end{aligned}$$

If some elements of $X$ are set by intervention, then the appropriate subset of the potentials and the modified normalizing constant must be used for these rows.

An appealing parameterization of the graphical model is with interventional potential functions of the form

$$\phi(x_i | x_{\pi(i)}, \theta) = \exp(b_{i,x_i} + \sum_{e \in \{<i,j>: j \in \pi(i)\}} w_{x_i, x_j, e}),$$

where each node $i$ has a scalar bias $b_{i,s}$ for each discrete state $s$, and each edge $e$ has a weight $w_{s_1,s_2,e}$ for each state combination $s_1$ and $s_2$ (so $\theta$ is the union of all $b_{i,s}$ and $w_{s_1,s_2,e}$ values). The gradient of the negative log-likelihood with potentials in this form can be expressed in terms of the training frequencies and marginal probabilities as

$$\begin{aligned} -\nabla_{b_{i,s}} \log p(X|\theta) = &- \sum_{d=1}^{m} \mathbb{I}_s(X_{d,i}) \\ &+ m\, p(x_i = s|\theta), \\ -\nabla_{w_{s_1,s_2,e}} \log p(X|\theta) = &- \sum_{d=1}^{m} \mathbb{I}_{s_1}(X_{d,i}) \mathbb{I}_{s_2}(X_{d,j}) \\ &+ m\, p(x_i = s_1, x_j = s_2|\theta). \end{aligned}$$

Under this parameterization the joint distribution is in an exponential family form, implying that the negative log-likelihood is a convex function in $b$ and $w$. However, the exponential family representation will have too many parameters to be identified from observational data. For example, with observational data we cannot uniquely determine the parameters in a directed 2-cycle. Even with interventional data the parameters remain unidentifiable because, for example,

[4] Note that in general the entire Markov blanket is needed to form the conditional distribution of a node $i$. In particular, the 'pseudo-Gibbs' sampler where we loop through the nodes in some order and sample $x_i \propto \phi(x_i | x_{\pi(i)})$ will not necessarily yield the appropriate stationary distribution unless certain symmetry conditions are satisfied (see Heckerman et al. (2000); Lauritzen and Richardson (2002)).

re-scaling an individual potential function does not change the likelihood.

To make the parameters identifiable in our experiments, we perform MAP estimation with a small $\ell_2$-regularizer added to the negative log-likelihood, transforming parameter estimation into a strictly convex optimization problem (this regularization also addresses the problem that the unique infimum of the negative log-likelihood may only be obtained with an infinite value of some parameters, such as when we have deterministic dependencies). Specifically, we consider the penalized log-likelihood

$$\min_{\theta} - \log p(X|\theta) + \lambda_2 ||\theta||_2^2,$$

where $\lambda_2$ controls the scale of the regularization strength.

The dominant cost of parameter estimation is the calculation of the node and edge marginals in the gradient. For parameter estimation in models where inference is not tractable, the interventional potential representation suggests implementing a pseudo-likelihood approximation (Besag, 1975). Alternately, an approximate inference method could be used to compute approximate marginals.

## 9 Convex Relaxation of Structure Learning

In many applications we may not know the appropriate graph structure. One way to write the problem of simultaneously estimating the parameters and graph structure is with a cardinality penalty on the number of edges $\mathcal{E}(G)$ for a graph structure $G$, leading to the optimization problem

$$\min_{\theta, G} - \log p(X|\theta) + \lambda \mathcal{E}(G),$$

where $\lambda$ controls the strength of the penalty on the number of edges. We can relax the discontinuous cardinality penalty (and avoid searching over graph structures) by replacing the second term with a group $\ell_1$-regularizer (Yuan and Lin, 2006) on appropriate elements of $w$ (each group is the elements $w_{\cdot,\cdot,e}$ associated with edge $e$), giving the problem

$$\min_{\theta} - \log p(X|\theta) + \lambda \sum_{e} ||w_{\cdot,\cdot,e}||_2, \qquad (1)$$

where $\theta = \{b_{\cdot,\cdot}, w_{\cdot,\cdot,\cdot}\}$. Solving this continuous optimization problem for sufficiently large $\lambda$ yields a sparse structure, since setting all values $w_{\cdot,\cdot,e}$ to zero for a particular edge $e$ is equivalent to removing the edge from the graph. This type of approach to simultaneous parameter and structure learning has previously been explored for undirected graphs (see Lee et al., 2006; Schmidt et al., 2008), but can not be used directly for DAG models (unless we restrict ourselves to a fixed node ordering) because of the acyclicity constraint.

Similar to Schmidt et al. (2008), we can convert the continuous and unconstrained but non-differentiable problem (1) into a differentiable problem with second-order cone constraints

$$\min_{\theta} - \log p(X|\theta) + \lambda \sum_{e} \alpha_e,$$
$$s.t. \quad \alpha_e \geq ||w_{\cdot,\cdot,e}||_2.$$

Since the edge parameters form disjoint sets, projection onto these constraints is a trivial computation (Boyd and Vandenberghe, 2004, Exercise 8.3(c)), and the optimization problem can be efficiently solved using a limited-memory projected quasi-Newton method (Schmidt et al., 2009)

## 10 Experiments

We compared the performance of several different graphical model representations on two data sets. The particular models we compared were:

- DAG: A directed acyclic graphical model trained with group $\ell_1$-regularization on the edges. We used the projected quasi-Newton method of Schmidt et al. (2009) to optimize the criteria for a given ordering, and used the dynamic programming algorithm of Silander and Myllymaki (2006) to minimize the regularized negative log-likelihood over all possible node orderings. The interventions are modeled as described in §5.
- UG-observe: An undirected graphical model trained with group $\ell_1$-regularization on the edges that treats the data as if it was purely observational. Specifically, it seeks to maximize $p(x_1, \ldots, x_n)$ over the training examples (subject to the regularization) and ignores that some of the nodes were set by intervention.
- UG-condition: An undirected graphical model trained with group $\ell_1$-regularization on the edges that conditions on nodes set by intervention. Specifically, it seeks to maximize $p(x_1, \ldots, x_n)$ over the training examples (subject to the regularization) on observational samples, and seeks to maximize $p(x_{-k}|x_k)$ over the training examples (subject to the regularization) on interventional samples where node $k$ was set by intervention.
- DCG: The proposed directed cyclic graphical model trained with group $\ell_1$-regularization, modeling the interventions as described in §5.

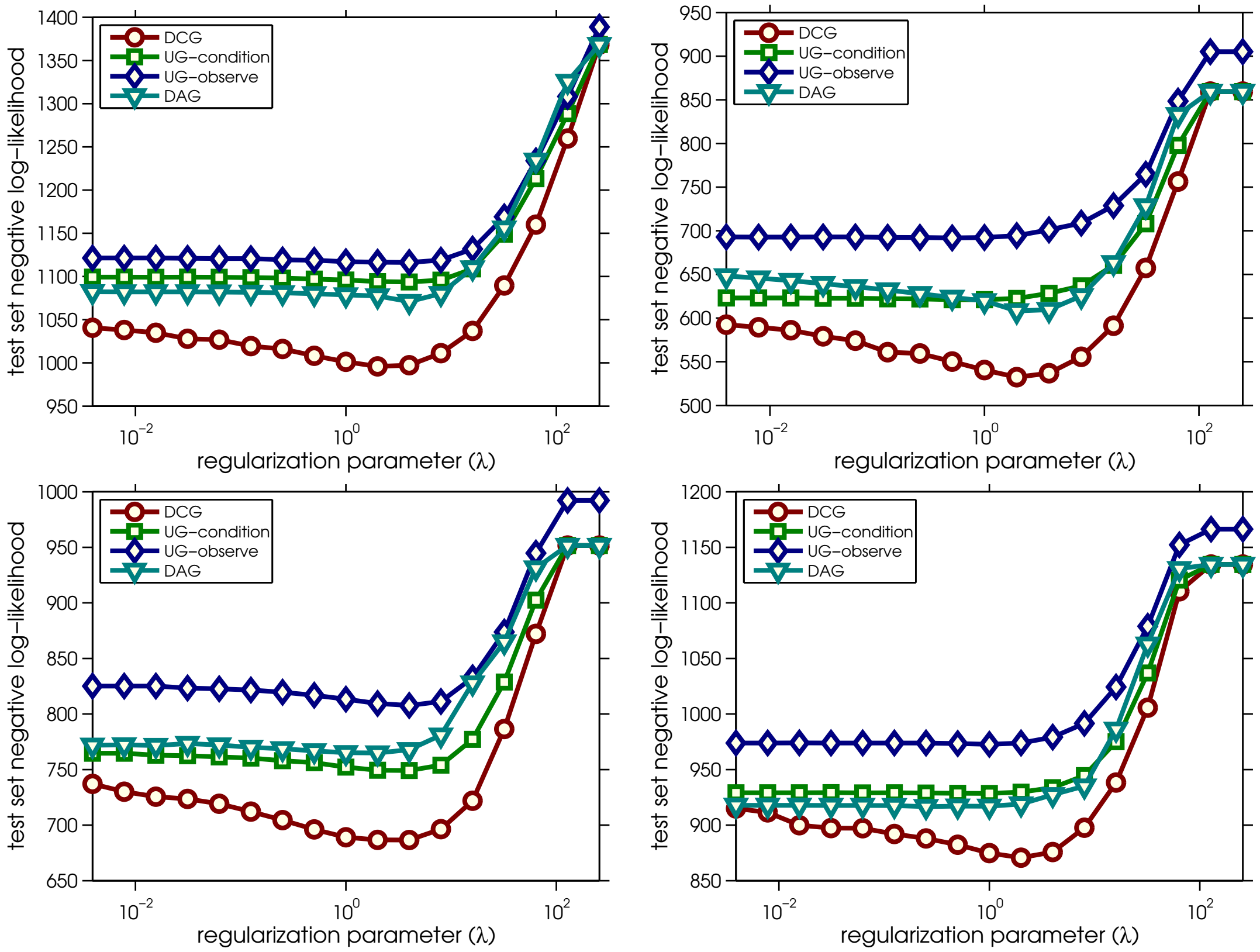

Figure 4: Results on data generated from 4 different DCG models.

To make the comparisons fair, we used a linear exponential family representation for all models. Specifically, the DAG model uses conditional probabilities of the form

$$p(x_i|x_{\pi(i)},\theta) = \frac{1}{Z_i}\exp(b_{i,x_i} + \sum_{j\in\pi(i)} w_{x_i,x_j,e}),$$

(where $Z_i$ normalizes locally), while the UG models use a distribution of the form

$$p(x_1,\ldots,x_n|\theta) = \frac{1}{Z}\exp(\sum_{i=1}^{n} b_{i,x_i} + \sum_{e:\{<i,j>\in\mathcal{E}\}} w_{x_i,x_j,e}),$$

and the DCG model uses the interventional potential functions described in §8.

To ensure identifiability of all model parameters and increase numerical stability, we applied $\ell_2$-regularization to the parameters of all models. We set the scale $\lambda_2$ of the $\ell_2$-regularization parameter to $10^{-4}$, but our experiments were not particularly sensitive to this choice. The groups used in all methods were simply the set of parameters associated with an individual edge.

### 10.1 Directed Cyclic Data

We first compared the performance in terms of test-set negative log-likelihood on data generated from an interventional potential model. We generated the graph structure by including each possible directed edge with probability 0.5, and sampled the node and edge parameters from a standard normal distribution, $\mathcal{N}(0,1)$. We generated 1000 samples from a 10-node binary model, where in 1/11 of the samples we generated a purely observational sample, and in 10/11 of the samples we randomly choose one of the ten nodes and set it by intervention. We repeated this 10 times to generate 10 different graph structures and parameterizations, and for each of these we trained on the first 500 samples and tested on the remaining 500. In Figure 4, we plot the test set negative log-likelihood (of nodes not set by intervention) against the strength of the group $\ell_1$-regularization parameter for the first 4 of these trials (the others yielded qualitatively similar results).

We first contrast the performance of the DAG model (which models the effects of intervention but can not model cycles) with the UG-condition method (which does not model the effects of intervention but can model cycles). In our experiments, neither of these methods dominated the other; in most distributions the optimal DAG had a small advantage over UG-condition for suitably chosen values of the regularization parameter, while in other experiments the UG-condition model offered a small advantage. In contrast, the DCG model (which models the effects of interventions and also allows cycles) outperformed both the DAG and the UG methods over all 10 experiments. Finally, in terms of the two UG models, conditioning on the interventions during training strictly dominated treating the data as observational, and in all cases the UG-observe model was the worst among the 4 methods.

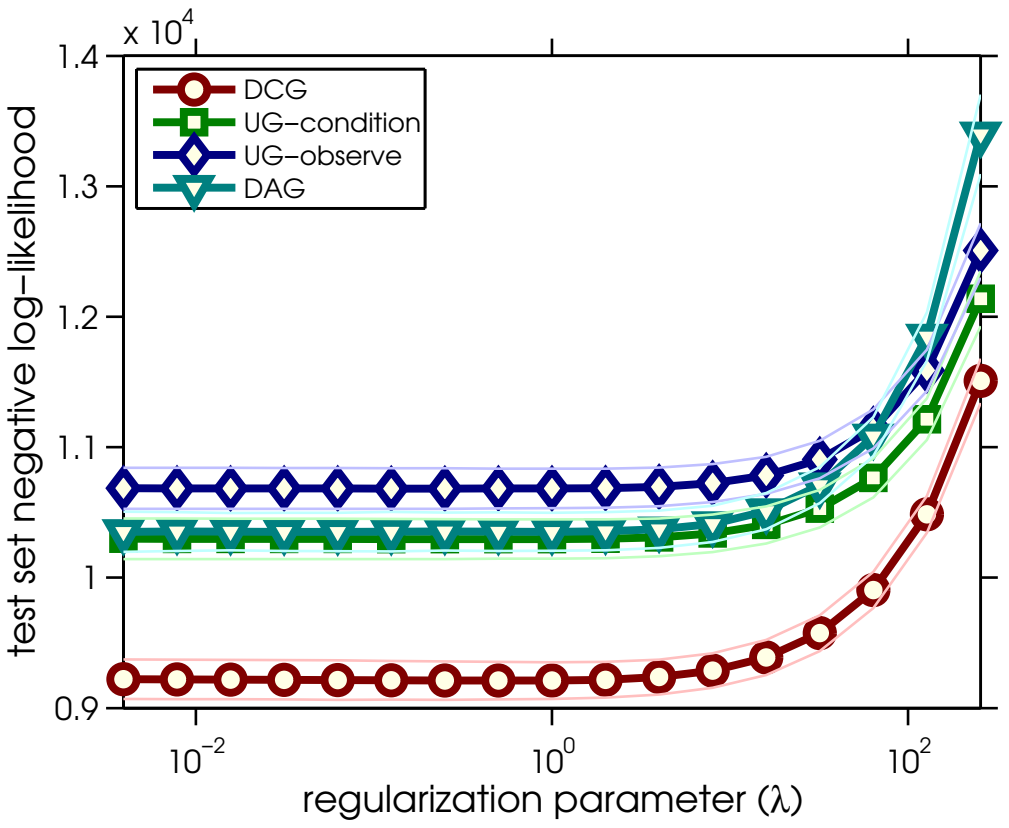

Figure 5: Mean results (plus/minus two standard deviations) on the expression data from Sachs et al. (2005) over 10 training/test splits.

### 10.2 Cell Signaling Network

We next applied the methods to the data studied in Sachs et al. (2005). In this study, intracellular multivariate flow cytometry was used to simultaneously measure the expression levels of 11 phophorylated proteins and phospholipid components in individual primary human immune system cells under 9 different stimulatory/inhibitory conditions. The data produced in this work is particularly amenable to statistical analysis of the underlying system, because intracelleular mulitvariate flow cytomery allows simultaneous measurement of multiple proteins states in individual cells, yielding hundreds of data points for each interventional scenario. In Sachs et al. (2005), a multiple restart simulated annealing method was used to search the space of DAGs, and the final graph structure was produced by averaging over a set of the most high-scoring networks. Although this method correctly identified many edges that are well established in the literature, the method also missed three well-established connections. The authors hypothesized that the acyclicity constraint may be the reason that the edges were missed, since they could have introduced directed cycles (Sachs et al., 2005). In principle, these edges could be discovered using DAG models of time-series data. However, current technology does not allow collection of this type of data (Sachs et al., 2005), motivating the need to examine cyclic models of interventional data.

In our experiments, we used the targets of intervention and 3-state discretization strategy (into 'under-expressed', 'baseline', and 'over-expressed') of Sachs et al. (2005). We trained on 2700 randomly chosen samples and tested on the remaining 2700, and repeated this on 10 other random splits to assess the variability of the results. Figure 5 plots the mean test set likelihood (and two standard deviations) across the 10 trials for the different methods. On this real data set, we see similar trends to the synthetic data sets. In particular, the UG-observe model is again the worst, the UG-condition and DAG models have similar performance, while the DCG model again dominates both DAG and UG methods.

Despite its improvement in predictive performance, the learned DCG graph structures are less interpretable than previous models. In particular, the graphs contain a large number of edges even for small values of the regularization parameter (this may be due to the use of the linear parameterization). The graphs also include many directed cycles, and for larger values of $\lambda$ incorporate colliders whose parents do not share an edge. It is interesting that the (pairwise) potentials learned for 2-cycles in the DCG model were often very asymmetric.

## 11 Discussion

While we have assumed that the interventions are 'perfect', in many cases it might be more appropriate to use DCG models with 'imperfect', 'soft', or 'uncertain' interventions (see Eaton and Murphy (2007)). We have also assumed that each edge is affected asymmetrically by intervention, while we could also consider undirected edges that are affected symmetrically. For example, we could consider 'stable' undirected edges that represent fundamentally associative relationships that remain after intervention on either target. Alternately, we could consider 'unstable' undirected edges that are removed after intervention on either target (this would be appropriate in the case of a hidden common cause).

To summarize, the main contribution of this work is a model for interventional data that allows cycles. It is therefore advantageous over undirected graphical models since it offers the possibility to distinguish between 'seeing' and 'doing'. However, unlike DAG mod-els that offer the same possibility, it allows cycles in the model and thus may be more well-suited for data sets like those generated from biological systems which often have natural cyclic behaviour.

## Acknowledgements

We would like to thanks the anonymous reviewers for helpful suggestions that improved the paper.